\documentclass{article}

\usepackage[dblblindworkshop, final]{neurips_2025}
\workshoptitle{Ai4 Science}

\usepackage[utf8]{inputenc} 
\usepackage[T1]{fontenc}    
\usepackage{hyperref}       
\usepackage{url}            
\usepackage{booktabs}       
\usepackage{amsfonts}       
\usepackage{nicefrac}       
\usepackage{microtype}      
\usepackage{xcolor}         
\usepackage{tabularx}
\usepackage[table]{xcolor}
\usepackage{pifont} 
\title{Enginuity: Building an Open Multi-Domain Dataset of Complex Engineering Diagrams}

\author{%
  Ethan Seefried\\
  Oak Ridge National Laboratory\\
  \texttt{seefriedej@ornl.gov}
  \And
  Prahitha Movva\\
  Independent Researcher\\
  \texttt{prahitha.movva03@gmail.com}
  \AND
  Naga Harshita Marupaka\\
  Independent Researcher\\
  \texttt{nagaharshitamarupaka@gmail.com}
  \And
  Tilak Kasturi\\
  Predii\\
  \texttt{tilak@predii.com}
  \And
  Tirthankar Ghosal\\
  Oak Ridge National Laboratory\\
  \texttt{ghosalt@ornl.gov}
}

\begin{document}

\maketitle

We propose \textit{Enginuity} - the first open, large-scale, multi-domain engineering diagram dataset with comprehensive structural annotations designed for automated diagram parsing. By capturing hierarchical component relationships, connections, and semantic elements across diverse engineering domains, our proposed dataset would enable multimodal large language models to address critical downstream tasks including structured diagram parsing, cross-modal information retrieval, and AI-assisted engineering simulation. Enginuity would be transformative for AI for Scientific Discovery by enabling artificial intelligence systems to comprehend and manipulate the visual-structural knowledge embedded in engineering diagrams, breaking down a fundamental barrier that currently prevents AI from fully participating in scientific workflows where diagram interpretation, technical drawing analysis, and visual reasoning are essential for hypothesis generation, experimental design, and discovery.

\section{Dataset Rationale}
Engineering diagrams encode the core knowledge of scientific and technical disciplines but remain inaccessible to AI (refer Appendix \ref{app: intro}) due to proprietary reasons. While current methods achieve 85\%+ accuracy on symbol detection, they struggle with relationship extraction—the critical bottleneck where performance drops by 25\%+ \cite{Jan_Marius_2025}, preventing true diagram understanding. No public dataset exists with >10K real-world engineering diagrams containing both component and structural relationship annotations. This gap prevents AI from participating in scientific workflows requiring visual-structural reasoning and system-level comprehension.

We propose Enginuity—50K annotated engineering diagrams starting with automotive domain (will be later expanded to include other domains) through our partnership with \textit{Predii}, an automotive AI company processing 2B+ repair jobs monthly. Automotive diagrams are ideal: they combine visual structure, text, and functional knowledge in exploded parts diagrams used by technicians globally. Our dataset will enable three core AI tasks: \textbf{(1) component detection (2) relationship extraction and (3) diagram VQA.} This will culminate in a CVPR 2026 workshop and shared task, and later into a leaderboard \textit{arena} to test capabilities of frontier AI models.

\section{Dataset Building Strategy}
To balance openness with real-world relevance, for data collection, we will adopt a two-pronged strategy:

\textbf{Public-domain automotive diagrams} – We will collect and annotate diagrams from declassified government vehicles and older vehicles in the public domain. These resources include both exploded parts diagrams and associated technical procedure manuals that explicitly reference the diagrams in repair workflows. Our industry collaborator will bring in domain experts for the human-labeling.

\textbf{Industry engagement framework:} We will establish a framework for private industry contributors (e.g., OEMs) to contribute “older vehicle parts diagrams” (5-15 years old) without disclosing proprietary information via our industry collaborator. This creates a pathway for researchers to access diverse, realistic datasets while giving industry partners a mechanism to engage with the research community.

Enginuity will contain 50K+ diagrams spanning powertrains, chassis, and body components from 500+ vehicle models. Annotations include hierarchical component relationships, spatial connections, part numbers, and functional roles, standardized to ISO/IEEE ontologies. Our 4-stage annotation pipeline uses AI for initial detection, dedicated teams for refinement, expert validation on 10\% samples, and active learning to reduce costs by 65\%.

Please refer to Appendices \ref{app: compare-dataset}, ~\ref{app:data-creation-pathway}, ~\ref{app:data-annotation}, \ref{app: data-part}, and \ref{app: data-kaggle} for further details.

\section{Defining the AI Task} 
Understanding and reasoning on scientific/engineering diagrams is a \textit{long-studied, difficult computer vision problem} \cite{Kembhavi2016DiagramWorth, Mani2020Digitization, montalvo1985diagram}, which requires progress in perception, structured representation, and multimodal reasoning. We believe no single group can solve this in isolation; it demands a sustained community effort supported by a shared benchmark dataset and leaderboard. \textbf{Primary task: parsing diagrams into structured graphs linking visual components to part identifiers.} We have a set of concrete subtasks spanning\textit{ component and symbol recognition, relationship extraction, functional context interpretation, diagram question answering, multimodal information retrieval, diagram-to-digital-twin alignment} (we elaborate more on these tasks in Appendix~\ref{app:tasks} and evaluation metrics in \ref{app: evals}.)

\section{Potential Acceleration}

\textit{Enginuity} will help to train foundation models for several hard downstream tasks including the ones mentioned in the previous section. The ability to automatically parse and reason over diagrams opens pathways to \textit{scientific acceleration}: \textbf{Algorithmic exploration of designs, Cross-domain transfer} - models trained on automotive diagrams generalize to mechanical/process engineering, \textbf{Digital twin generation}—automated conversion from 2D diagrams to 3D simulations, and \textbf{Knowledge preservation \& harmonization} across notation changes spanning decades. Practical \textit{downstream accelerations} include: \textbf{Design optimization, Simulation integration, and Automated documentation}. Please refer to Appendix \ref{app: potential} for further details.

By making such a resource openly available, we will \textbf{lower the entry barrier for researchers, catalyze cross-disciplinary innovation, and unite the AI and scientific communities} to tackle one of the most challenging problems in automated scientific discovery. The Enginuity dataset would enable frontier AI models to finally crack the spatial reasoning bottleneck in engineering diagram comprehension by providing the first large-scale (50K+), multi-domain training corpus with rich spatial annotations that bridges the critical gap between isolated symbol recognition and true system-level understanding. This breakthrough would transform AI from merely detecting components to actually comprehending how complex engineering systems interconnect and function, unlocking automated digital twin generation, cross-domain technical reasoning, and real-time engineering design validation that could accelerate the entire Industry 4.0 transformation\footnote{\url{https://www.mckinsey.com/featured-insights/mckinsey-explainers/what-are-industry-4-0-the-fourth-industrial-revolution-and-4ir}}.



\bibliographystyle{unsrtnat}
\bibliography{bib}


\appendix

\section{Supplementary Material}

\subsection{Introduction}
\label{app: intro}
Engineering diagrams serve as the universal visual language of scientific and technical disciplines, encoding critical knowledge about system architectures, process flows, circuit designs, molecular structures, and experimental setups that form the backbone of scientific research and engineering practice. These diagrams represent decades of accumulated scientific knowledge in a structured, standardized format that researchers rely on for design, analysis, communication, and innovation in different fields of science and technology. In essence, these diagrams are blueprints and instruments of science \cite{bigg2016diagrams, moktefi2017diagrams, perini2013diagrams}, ensuring that the components of an experiment fit and function together, reducing costly design errors and accelerating assembly. However, due to the inherent complexity of these diagrams, even the frontier AI models struggle to comprehend them \footnote{\url{https://www.businesswaretech.com/blog/benchmarking-ai-on-tables-and-engineering-drawings-results-findings}}. To train modern AI applications, these complex diagrams needs human labeling. However, most such diagrams remain locked in proprietary silos, out of reach for researchers. The lack of a large, open, expert-labeled, cross-disciplinary dataset of complex engineering diagrams is a major bottleneck \cite{beutenmuller2023topological}, limiting the ability of AI to generalize, generate, or link diagrams to digital twins.

Hence, we propose \textit{Enginuity}--a large-scale, open, multi-domain dataset of complex engineering diagrams to bridge the gap and help train our models for this hard task. Additionally, we want to catalyze breakthroughs in AI-assisted knowledge extraction from complex diagrams, multimodal information retrieval, scientific reasoning, complex diagram parsing, and design generation by building a community around the dataset and uniting researchers across disciplines. This initiative will culminate in a shared task challenge and workshop that we are already planning for CVPR 2026. If successful, we would like to build a \textit{arena} in lines of \textit{lmsys}\footnote{\url{https://lmsys.org/}} where the frontier AI models will be tested on this hard benchmark. With the assistance of our automotive industry collaborator (also coauthor of this proposal) who is bringing in the wealth of domain knowledge of automotive diagrams and data, we are confident to launch this initiative; although automotive diagrams will form a significant portion of Enginuity 1.0, but we are not only limited to it. \textbf{Why automotive diagrams matter?} Automotive vehicles provide a uniquely compelling case study. Modern repair workflows are highly dependent on exploded parts diagrams, which allow technicians to identify replacement parts by linking a visual diagram to a set of associated components, assemblies, one-time-use parts, and manufacturer-specific part numbers. In practice, technicians often use natural language queries (e.g., “front-left brake caliper”) to navigate these diagrams, bridging between visual structure and symbolic identifiers. The tight coupling of visual, textual, and functional knowledge mirrors challenges across many scientific domains, making automotive diagrams an ideal testbed for advancing AI methods in multi-modal scientific reasoning \cite{boon2008diagrammatic}.

\subsection{Comparison to Existing Datasets}
\label{app: compare-dataset}
\begin{itemize}
    \item \textbf{Small or narrow in scope} – Current datasets focus on limited domains such as P\&ID, electrical schematics, or specific CAD formats.
    \item \textbf{Proprietary or restricted} – Many high-value diagrams remain locked within corporate, government, or private archives, inaccessible due to IP, competition, or regulation.
    \item \textbf{Domain-locked} – Public datasets rarely span the full breadth of scientific diagrams, leaving cross-disciplinary AI models underdeveloped. \cite{khan2024fine}
    \item \textbf{Fragmented \& siloed} – Data is often trapped in discipline-specific formats and communities, with no interoperability or shared standards.
    \item \textbf{Legacy diagram barriers} – Decades of archival diagrams use shifting notations and conventions, hindering knowledge transfer between veteran experts and newer technologists. Future models trained on this dataset could harmonize visual languages and bridge generational gaps.
\end{itemize}

Current datasets are too narrow to support the proposed accelerations:
\begin{itemize}
    \item \textbf{Handwritten engineering diagrams (circuits)} \cite{johannes_bayer_2025} – Focused exclusively on small-scale electrical circuits, lacking diversity in both domain and diagram type.
    \item \textbf{SiED: Symbol Classification Engineering Diagrams} \cite{Elyan2020SiED} – Provides isolated symbol classification but does not address higher-order structure, system-level relationships, or domain generalization.
\end{itemize}

By contrast, our dataset is \textbf{multi-domain, multi-scale, and system-focused}, capturing the full spectrum from individual components to assemblies and process flows. This breadth directly enables generalizable AI models capable of supporting scientific discovery, industrial design, and digital twin development in a way that current resources cannot.

\subsection{Data Creation Pathway}
\label{app:data-creation-pathway}
The dataset will be sourced through collaborations with automotive industry partners, providing technical schematics and parts lists across diverse car models and years. These materials will span thousands of diagrams covering powertrains, chassis, electronics, and body components.

Annotation will combine semi-automated preprocessing with expert review, leveraging (\textit{Anonymous}) AI platform to align technical manuals with exploded parts diagrams. Human-in-the-loop annotation by technicians will ensure accuracy and domain fidelity.

Key annotation targets include: 
\begin{itemize}
    \item Part–number relationships linking visual components to identifiers.
    \item Hierarchy tagging of components, sub-components, assemblies, and one-time-use parts.
    \item Specifications and usage types for functional context.
    \item System–subsystem clustering of diagrams for structured retrieval.
    \item Diagram question answering to support multi-modal reasoning tasks.
\end{itemize}

This approach ensures the dataset is both scalable and high-quality, directly reflecting real-world automotive repair workflows. 

\textbf{Proposed dataset features:}
\begin{itemize}
    \item \textbf{Scale} - More than 50K diagrams from automotive systems (\textit{there are over 50,000 distinct vehicle year–make–model–engine combinations in the North American passenger vehicle market alone}), and process engineering.
   \item \textbf{Focus} - Emphasis on physical structure and relationships; excluding electrical schematics.
    \item \textbf{Intra-domain diversity} - Coverage from prototypes to production designs, and from small parts (e.g., gears) to full systems (e.g., drivetrains).
    \item \textbf{Rich metadata \& labels} - Domain, Component types, Connections, and Functional roles.
    \item \textbf{Multi-source availability} - Drawn from public archives, industrial standards, and selectively shared industry diagrams.
\end{itemize}

\subsection{Annotation Strategy and Pipeline}
\label{app:data-annotation}
A sophisticated, multi-stage human-in-the-loop (HITL) pipeline will ensure both scalability and accuracy while keeping costs feasible. Our approach integrates \textit{Anonymous's} AI platform with active learning to progressively improve annotation quality and efficiency. Funding will directly support the fine-grained annotation and preparation of training, validation, and test sets for the challenge and, more broadly, the long-term academic–industry infrastructure needed to move the field forward.

\textbf{Stage 1: AI-driven preprocessing.} \textit{Anonymous's} domain-specialized LLMs and vector embeddings will act as a machine labeler, performing an initial pass to detect lines, arrows, text regions, and component clusters. Heterogeneous input formats (PDF, DXF, SVG, raster scans) will be standardized into a normalized digital format, with consistent units and vectorized representations for downstream processing.

\textbf{Stage 2: Dedicated refinement.} Low-complexity tasks such as bounding box adjustments, OCR text verification, and alignment of auto-detected components will be handled by a dedicated annotation team in collaboration with our industry partner. This approach ensures consistent quality control, leverages domain-trained annotators, and frees expert scientists to focus on higher-value annotations.

\textbf{Stage 3: Expert annotation.} A domain expert team (e.g., technicians, engineers) will create a “golden set” and review 5–10\% of annotations for complex components, assemblies, and functional roles. Their input will seed the validation process and establish quality benchmarks.

\textbf{Stage 4: Active learning loop.} Validated annotations will bootstrap model training. Updated models will then auto-label subsequent batches, focusing expert/crowd review on uncertain or novel cases. This iterative process continuously reduces annotation cost while improving model accuracy.

\textbf{Annotation schema.} Labels extend beyond component metadata to include:
\begin{itemize}
    \item Object segmentation and bounding boxes
    \item Attributes (e.g., component type, specifications, usage)
    \item Relationship and topology graphs (connections, spatial orientation)
    \item Functional and hierarchical structures (system $\rightarrow$ subsystem $\rightarrow$ component)
    \item Temporal metadata for standards evolution and difficulty scoring
\end{itemize}

\textbf{Standardization and Ontology Alignment.} All labels will be mapped to or extend existing engineering ontologies (e.g., IEEE, ISO, ISA) to ensure interoperability and reusability across domains.

This tiered pipeline balances automation, crowdsourcing, and expert input, providing a technically feasible and cost-efficient path to generate rich, multi-level annotations at scale.

\subsection{Data Partitioning}
\label{app: data-part}
A key element of the pipeline is the dataset partitioning strategy. The annotated corpus will be divided into four subsets: training, validation, testing, and a withheld testing set reserved for the organized competition. This four-way split is intentional: the training and validation sets support model development and hyperparameter tuning, while the first testing set allows for transparent baseline reporting. The withheld test set, by contrast, is unseen during model development and will be used to score submitted competition entries. This design prevents data leakage and overfitting, ensuring a fair and reproducible evaluation protocol.  

To assess generalizability, the withheld test set will additionally include diagrams sourced from private organizations outside the automotive domain. These diagrams, which may reflect distinct conventions in engineering drawing styles, annotation symbols, or system complexity, introduce a domain shift. Incorporating such heterogeneous material ensures that models are not merely optimized for a narrow distribution of automotive schematics but can extend to scientific and engineering diagrams more broadly. This robustness check is essential for evaluating real-world applicability, where trained systems must handle variability in style, notation, and structure.

\subsection{Task Suite Details} 
\label{app:tasks}
\begin{itemize}
    \item \textbf{Component and Symbol Recognition} – Detecting and classifying parts, symbols, and visual primitives across heterogeneous diagram styles.
    \item \textbf{Relationship Extraction} – Inferring spatial and logical connections between components to construct a machine-readable graph representation.
    \item \textbf{Functional Context Interpretation} – Reasoning about the role of components and subsystems (e.g., identifying assemblies, one-time-use parts, or failure-prone connections) in order to understand the diagram’s operational purpose.
    \item \textbf{Diagram Question Answering (DQA)} – Enabling natural language queries over diagrams (e.g., “Which components must be removed before replacing the brake caliper?”), requiring joint reasoning across visual, symbolic, and textual modalities.
    \item \textbf{Diagram-to-Digital-Twin Alignment} – Mapping diagrams into structured formats compatible with digital twin models for simulation, retrieval, and knowledge transfer.
\end{itemize}

By explicitly defining these tasks, the dataset supports a hierarchical research agenda: from low-level perception (symbol detection) to mid-level structure induction (relationship extraction) to high-level reasoning (functional interpretation and question answering). In doing so, it creates a testbed not only for advancing multimodal AI but also for probing the limits of scientific reasoning from visual artifacts.

\subsection{Evaluation Metrics} 
\label{app: evals}
To ensure rigor and comparability, we will define clear baseline metrics for each core task. For symbol and component recognition, we propose to use \textbf{Mean Average Precision (mAP)} at standard IoU thresholds, following established conventions in object detection \cite{zou2023object}. For relationship extraction and diagram graph construction, we propose a \textbf{graph accuracy metric}, computed as the proportion of correctly predicted edges and node labels compared to ground-truth graphs \cite{alenazi2015comprehensive}.

These metrics are intended as initial baselines; we will refine and extend them in consultation with the research community, particularly for higher-level tasks such as diagram question answering (DQA) and diagram-to-digital-twin alignment, where specialized evaluation protocols may be needed and already some established baseline metrics are available to start with. This approach balances concreteness with flexibility, ensuring fair benchmarking while leaving room for iteration as the shared task evolves.

\subsection{File Formatting}
\label{app: file-format}
Finally, a deliberate effort will be made to address the challenge of file format inconsistency. In industrial contexts, schematics are frequently distributed as PDFs, often exported from proprietary CAD tools or drawing software. These PDFs vary in resolution, compression, layering, and embedded metadata, making them difficult to parse systematically \cite{yang2023comprehensiveendtoendcomputervision}. Some may contain vector graphics, while others are rasterized scans of paper documents, introducing noise and heterogeneity. Left unaddressed, such inconsistency can bias downstream models and hinder reproducibility. To mitigate this, all diagrams will be converted into a standardized, machine-readable digital format (e.g., high-resolution vector or normalized raster images). By enforcing a uniform representation, we establish consistency across the dataset, enabling reliable annotation, training, and benchmarking. Moreover, this standardization ensures that the resource can be easily extended with new contributions in the future, without inheriting the ad hoc variability of industrial documentation practices.  

\subsection{Dataset Availability}
\label{app: data-kaggle}
The curated dataset will be released under an open license and made fully publicly accessible through \textbf{Kaggle}, which provides a robust infrastructure for large-scale dataset hosting, versioning, and community engagement. Each release will be accompanied by a detailed \textbf{Kaggle Datacard} documenting data provenance, annotation schema, licensing terms, and known limitations, in line with best practices for responsible dataset publication.

To foster reproducibility and accelerate adoption, we will also provide:
\begin{itemize}
    \item \textbf{Baseline models and benchmarks} implemented in common frameworks (PyTorch) with evaluation scripts.
    \item \textbf{Leaderboards} hosted on Kaggle for tracking progress during the organized CVPR competition and beyond.
    \item \textbf{Tutorial notebooks and usage examples} illustrating key tasks such as component recognition, diagram retrieval, and question answering.
    \item \textbf{Community support channels}, including a discussion forum and issue tracker to encourage feedback, error reporting, and collaborative extensions.
\end{itemize}

Long-term, the dataset will be maintained with clear \textbf{versioning protocols} to ensure stability of benchmark splits while allowing for incremental expansion to new domains. This strategy ensures not only a fair competition environment but also a sustainable resource for the broader AI-for-science community.

\subsection{Acceleration Potential}
\label{app: potential}
The proposed dataset will significantly accelerate research at the intersection of computer vision, scientific reasoning, and engineering design. 
By providing a large-scale, multi-domain, and richly annotated resource, it enables a new class of scientific questions centered on the understanding, interpretation, and generation of engineering diagrams across multiple domains (\textit{automotive domain} is main focus for \textbf{Enginuity v1.0}, but will be expanded to other domains later). 
Unlike text or tabular data, diagrams encode structural and relational knowledge that directly underpins physical systems. Unlocking this information algorithmically accelerates both basic science and applied innovation. 
\subsection*{Scientific Acceleration}
\begin{itemize}
    \item \textbf{Algorithmic exploration of designs} – Models can propose, evaluate, and iterate design variations directly from diagrams, reducing the need for costly manual drafting cycles.
    \item \textbf{Cross-domain transfer} – Trained systems can generalize across automotive, mechanical, and process-engineering domains, enabling comparative studies and accelerating design in areas where labeled data is scarce.
    \item \textbf{Knowledge preservation \& harmonization} – AI models trained on diagrams spanning decades can bridge generational gaps in notation, style, and conventions, preserving institutional knowledge while making it accessible to new engineers and scientists.
\end{itemize}

\subsection*{Applied Acceleration}
Practical downstream accelerations include:
\begin{itemize}
    \item \textbf{Design optimization} – Tools that automatically suggest structural improvements or highlight inefficiencies in complex assemblies.
    \item \textbf{Simulation integration} – Data-driven conversion of diagrams into machine-readable representations suitable for digital twin simulations, improving prediction accuracy and reducing experimental overhead.
    \item \textbf{Automated documentation} – Generation of consistent metadata and structured component lists from legacy diagrams, enabling interoperability across industry and research archives.
\end{itemize}

\subsection{Budget and Timeline}
Our overall estimated budget for data collection \& annotations ($\$150$K), infrastructure support ($\$30$K), baselines \& evaluation ($\$20$K) would approximately amount to $\$200$K US dollars. By the end of month 12 we will have 50K annotated images on Hugging Face for the research community. If successful, we would anticipate further support from the industry to launch the \textit{arena}.

\subsection{Acknowledgments}

We thank Predii AI for supplying and helping curate the initial engineering diagrams and annotations used in this work. We also acknowledge the Spallation Neutron Source (SNS) at Oak Ridge National Laboratory for institutional support and resources that enabled this research.

\end{document}